\documentclass{article}

\usepackage{arxiv}

\usepackage[utf8]{inputenc} 
\usepackage[T1]{fontenc}    
\usepackage{hyperref}       
\usepackage{url}            
\usepackage{booktabs}       
\usepackage{amsfonts}       
\usepackage{microtype}      
\usepackage{lipsum}
\usepackage{graphicx}
\usepackage{amsmath}
\usepackage{tikz}
\usepackage{pgfplots}
\usepackage{tcolorbox}
\usepackage{float}
\usepackage{pgfplotstable}
\usepackage[skip=0pt]{caption}
\usepackage{array}
\usepackage{pifont}
\usepackage{enumitem}
\usepackage{titlesec}
\usepackage{subcaption}
\usepackage{fullpage}
\usepackage{authblk}

\pgfplotsset{compat=1.18} 
\graphicspath{ {./images/} }
 
\title{CBEval: A Framework for Evaluating and Interpreting Cognitive Biases in LLMs}

\author[1]{Ammar Shaikh}
\author[2]{Raj Abhijit Dandekar}
\author[2]{Sreedath Panat}
\author[2]{Rajat Dandekar}

\affil[1]{Independent researcher\\ \texttt{ammarshaikh1510@gmail.com}}
\affil[2]{Vizuara AI Labs \\ \texttt{raj@vizuara.com}, \texttt{sreedath@vizuara.com}, \texttt{rajat@vizuara.com}}

\begin{document}

\maketitle
\begin{abstract}
Rapid advancements in large language models (LLMs) has significantly enhanced their reasoning capabilities. Despite improved performance on benchmarks, LLMs exhibit notable gaps in their cognitive processes. Additionally, as reflections of human-generated data, these models have the potential to inherit cognitive biases, raising concerns about their reasoning and decision making capabilities. In this paper we present a framework to interpret, understand and provide insights into a host of cognitive biases in LLMs. Conducting our research on frontier language models, we're able to elucidate reasoning limitations and biases, and provide reasoning behind these biases by constructing influence graphs that identify phrases and words most likely responsible for biases manifested in LLMs. We further investigate biases such as round number bias and a cognitive bias barrier revealed when noting framing effects in language models.
\end{abstract}

\section{Introduction}
\label{sec:intro}
Large language models have emerged as powerful instruments that can automate reasoning process in a host of domains ranging from Olympiad level math problem solving \cite{trinh2024solving} to code generation \cite{chen2021evaluating} and financial planning \cite{yang2023fingpt}.
LLMs have seen widespread adoption among people to automate and refine tasks involving decision making, logical thinking and critical reasoning. \\
However, despite such promising results on numerous benchmarks, LLMs still possess surprising knowledge gaps caused due to a host of reasons varying from engineering heuristics such as tokenization \cite{wang2024tokenization} to limitations in the training data itself, such as data bias or a lack of up-to-date information. Furthermore, due to being trained on extensive data accumulated and refined by humans over the years, these LLMs present grounds for inherent cognitive bias as characterized in humans \cite{tversky1974judgment}. \\
While there have been promising demonstrations of GPT-based models on tasks involving creativity \cite{esser2024scaling} and critical thinking \cite{zhou2023language}, slight variations in input prompt requests can lead to vastly different output responses\cite{mohammadi2024wait}. Furthermore, due to being next token predictors scaled to a huge text, the world models formed by language models possess knowledge gaps unapparent on the surface level.  The presence of such knowledge gaps motivates us to consider various kinds of biases and how they may influence the model's behavior, potentially affecting its predictions, generalizations, and overall reliability across different tasks and domains. \\
Some of the primary works on investigating the reasoning process in humans focuses on evaluating various cognitive biases and how they affect human behavior and decision making \cite{kahneman2013prospect}. While sufficient study has been performed with regards to humans as respondents, LLMs still remain a relatively new subject for querying such biases.
Several recent works have explored cognitive bias and its influence on a language model's ability to think.
\cite{hagendorff2023human} Explain the presence of system 2 reasoning in GPT-4 grade models and how it's capabilities differ from earlier LLMs in specially designed cognitive reflective tests.
At the same time \cite{mirzadeh2024gsm} explores logical reasoning and understanding capabilities in GPT-4 and GPT-o1 grade models and argue for a lack of true reasoning in LLMs, by presenting modifications to the GSM-8K benchmarks and observing performance drops supporting their claim. 
\cite{mahowald2024dissociating} Discuss the correlation between language and cognitive reasoning and assess the symbiosis between thinking and linguistic competence.
\cite{shaki2023cognitive} Probe for cognitive effects in GPT-3 through a unique question answering mechanism to uncover biases such as priming effect, distance and size congruity effect.
\cite{binz2023using} Treat GPT-3 as a subject in cognitive psychology, and note how slight prompt variations can lead to subpar model responses.
\cite{abramski2023cognitive} Detect how LLMs exhibit an underlying "math anxiety" that diminishes with improvement in model capabilities.
Similarly, \cite{opedal2024language} explore cognitive biases that show up in LLM thinking process by navigating through word problems and arithmetic tasks while reflecting upon similarities to those uncovered in children. 
 \cite{wang2024will} focuses on representativeness heuristic bias in LLMs such as GPT-4 and Llama-2. The authors construct a dataset called REHEAT designed to test LLMs’ susceptibility to biases such as base rate and conjunction fallacy. 
Significant work also dwells upon the influence of downstream alignment processes such as RLFH on a model's behavior. \cite{mohammadi2024creativity}  emphasize on the lack of creativity in RLHF tuned LLM responses, and how they compare to base models, while \cite{itzhak2023instructed}  test cognitive biases in leading LLMs, whilst presenting evidence on higher presence of biases in RLHF aligned models as compared to base counterparts.

While previous works explore cognitive biases in humans\cite{kahneman2013prospect}, the continuous refinement of LLMs necessitates corresponding research on how these biases manifest in AI. Investigating LLMs through this lens may help uncover human-like limitations and implicit harmful tendencies, which are critical for AI safety and model alignment.
In this paper, we present a framework to interpret the presence of  emergent cognitive bias resulting from the training process in frontier large language models.
We believe our methods to defer from prior works (\cite{echterhoff2024cognitive}, \cite{jiang2024peek} in the interpretability it offers. While significant prior work has explored framing effect, anchoring effect(\cite{talboy2023challenging}, \cite{jones2022capturing}, \cite{suri2024large}) and availability heuristic \cite{wang2024will}, we build upon this by introducing round number bias, a bias that hasn’t been explored in LLMs yet.
In particular, we investigate language models for 5 cognitive biases: 
\begin{enumerate}[itemsep=0pt, parsep=0pt]
    \item Framing effect
    \item Anchoring effect
    \item Number bias
    \item Representativeness heuristic
    \item Priming effect
\end{enumerate}

We touch upon how traits particular, and often attributed as a strength to model training, such as strong in-context learning, or having a higher number of logical reasoning tasks in training data, influence these cognitive biases and shape model behavior.
Furthermore, we attempt to interpret the underlying reason for emergent cognitive biases by borrowing previously explored ideas in game theory through Shapley value analysis and noting influence graphs on model outputs \cite{mohammadi2024wait}. While significant progress has been made in applying Shapley values and SHAP to interpreting machine learning models, we believe our work to be novel in using Shapley score analysis as means to decipher and corroborate the underlying biases in cognitive reasoning exhibited by frontier LLMs. Through our research on framing effect, we're able to uncover the presence of a cognitive bias barrier, a limitation on the robustness in reasoning processes in large language models such as GPT-4o.\\
Our work aims to provide light into the underlying reasoning mechanisms in LLMs, and can be reproduced conveniently in both open and close sourced models while being cost effective.

\section{Methodology and Theoretical Framework}
\label{sec:theory}

Drawing on ideas emergent in game theory, we rely on Shapley score attribution to measure the significance of the input prompt on model response. \\
We implement our Shapley value analysis in a manner similar to \cite{mohammadi2024wait}, considering input words as players in a cooperative game to determine the output token probabilities. We template the entire prompt as consisting of context setting constants along with a set of words that are contributing players across all possible sets of coalitions. This practice of keeping some words as constants across all coalitions is done with thoughtful consideration in order to reduce the cost incurred in our experimentation which scale exponentially as a factor of number of variable words/players in coalition. \\
Considering the set $X = \{ x_1, x_2, \dots, x_n \}$ representing the players in a cooperative game with a payoff for the entire coalition, Shapley value estimates each player's contribution to this payoff, providing a solution where each player receives a reward proportional to their influence in the coalition.
For a player $x_i$ under the value function $v$, the Shapley value is denoted as $\phi_i(v)$ and is calculated using the following formula:
\begin{figure}[!h]
\begin{equation}
\phi_i(v) = \frac{1}{N!} \sum_{S \subseteq X \setminus \{x_i\}} |S|! (N - |S| - 1)! \left[ v(S \cup \{x_i\}) - v(S) \right]
\end{equation}
\end{figure}
That is, the Shapley value of player $x_i$ is the average of player $x_i$'s contribution to each coalition $\mathcal{S}$ weighted by $|\mathcal{S}|! (N - |\mathcal{S}| - 1)!$, the number of permutations in which the coalition can be formed. \\
Modelling this to reflect the output of a response by LLM as a function of input prompts, we can assess the importance of individual players in the input on producing the output. In all of our experiments involving Shapley score attribution, we consider individual words in the prompt as players in a cooperative game, influencing the probability of the output under consideration.

\begin{table}[H]
\centering
\caption{Cognitive Biases}
\label{tab:cognitive-biases}
\begin{tabular}{|p{0.18\textwidth}|p{0.42\textwidth}|p{0.40\textwidth}|}
\hline
\textbf{Cognitive Bias} & \textbf{Explanation} & \textbf{Example} \\
\hline
Framing effect & The way information is presented (framed) influences decision-making, even when the underlying information is identical. & When told a surgery has an 80\% survival rate, patients are more likely to choose it than when told it has a 20\% mortality rate. \\
\hline
Anchoring effect & The tendency to rely too heavily on an initial piece of information encountered (the anchor) when making decisions. & In salary negotiations, the first number mentioned often becomes the anchor, influencing the final agreed amount. \\
\hline
Round number bias & The tendency to prefer round numbers or interpret them as more significant in decision making process. & People are more likely to set goals or make estimates using round numbers (e.g., losing 10 pounds instead of 9 or 11). \\
\hline
Representativeness heuristic & Judging the probability of something based on how closely it resembles our mental prototype, rather than statistical likelihood. & Assuming a well-dressed, articulate person is more likely to be a CEO than a janitor, without considering the relative numbers of each profession. \\
\hline
Priming effect & When exposure to one stimulus influences the response to a subsequent stimulus, without conscious awareness. & Participants primed with words related to elderly stereotypes walked more slowly when leaving the experiment room. \\
\hline
\end{tabular}
\end{table}

While we test for the stated biases on all LLMs as shown in Table~\ref{tab:llm-evaluations}, we use GPT-4o and GPT-4o mini as the primary running examples throughout our discussions to dwell upon and explain these biases better in frontier LLMs.

\section{Implementation and Findings}
\label{sec:impl}

\subsection{Framing Effect}
\label{sec:frame}
\textit{Framing effect} is a cognitive bias where people react differently to a particular choice or situation depending on how it is presented or "framed." The same information can lead to different decisions based on whether it is framed in a positive or negative way. \\
In simpler terms, framing effect refers to the difference in perceptions caused by slight modifications in the presentation of the same underlying information.
In our experimentation, we start with an initially positively framed prompt and modify it by adding a negative connotation while maintaining the same underlying idea. In both cases, we analyze the influence of the words in prompt through Shapley value computation and use it to gain an understanding on how framing effect affects model decision making.

\begin{tcolorbox}[
    colback=white,
    colframe=black,
    boxrule=1pt,
    arc=5mm, 
    left=10pt,right=10pt,top=10pt,bottom=10pt,
    boxsep=0pt,
    width=\textwidth
]
\textit{Positive frame}: Given two stocks A and B, which stock do you invest in if stock B makes a profit 70\% of the time?\\[10pt]
\textit{Negative frame}: Given two stocks A and B, which stock do you invest in if stock B makes a loss 30\% of the time?
\label{fig:framing_effect_prob}
\end{tcolorbox}

Despite being presented differently, an underlying notion of equivalent success rate is maintained in both prompts.

First we present the favoured option for both sentences: 

\begin{minipage}{\textwidth}
\begin{figure}[H]
    \centering
    \begin{tikzpicture}
    \begin{axis}[
        name=plot1,
        ybar,
        bar width=15pt,  
        ylabel={Preference Probability},  
        symbolic x coords={Stock A, Stock B},
        xtick=data,
        ymin=0,
        ymax=1,  
        ytick={0,0.2,0.4,0.6,0.8,1},  
        nodes near coords,
        nodes near coords align={vertical},
        nodes near coords style={/pgf/number format/fixed, /pgf/number format/precision=4},  
        width=5cm,  
        height=4cm,   
        enlarge x limits=0.2,
        at={(0,0)},
        anchor=south west
    ]
    \addplot coordinates {(Stock A, 0.025) (Stock B, 0.9744)};
    \end{axis}
    
    \node[anchor=north, yshift=-5mm] at (plot1.south) {\textbf{Positive Frame}};  
    
    \begin{axis}[
        name=plot2,
        ybar,
        bar width=15pt,  
        ylabel={Preference Probability},  
        symbolic x coords={Stock A, Stock B},
        xtick=data,
        ymin=0,
        ymax=1,  
        ytick={0,0.2,0.4,0.6,0.8,1},  
        nodes near coords,
        nodes near coords align={vertical},
        nodes near coords style={/pgf/number format/fixed, /pgf/number format/precision=4},  
        width=5cm,  
        height=4cm,   
        enlarge x limits=0.2,
        at={(6.5cm,0)},  
        anchor=south west
    ]
    \addplot coordinates {(Stock A, 0.9131) (Stock B, 0.08682)};
    \end{axis}
    
    \node[anchor=north, yshift=-5mm] at (plot2.south) {\textbf{Negative Frame}};  
    \end{tikzpicture}
    \caption{Comparison of preference stock for positively and negatively framed prompts}
    \label{fig:stock-comparison}
\end{figure}
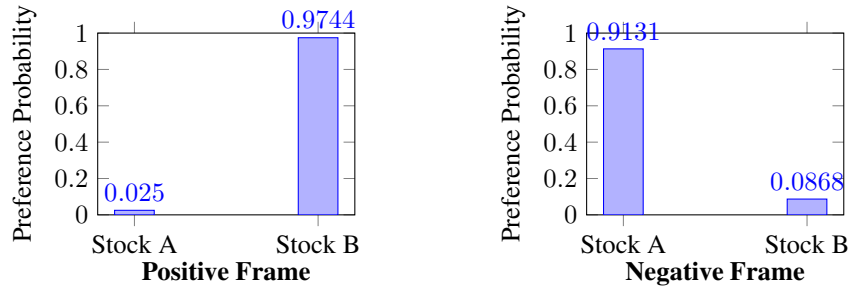
\end{minipage}

While the model exhibits a strong preference for stock B in prompt 1, the model's preferred output shifts significantly against stock B when the prompt is negatively framed, indicating a bias induced due to the manner the prompt is framed in. 
Note that we're not comparing whether the model's choice in either of the prompts is accurate to any degree or not, but simply evaluating for the change in output response due to the differences in framing of prompts.

Analyzing further, we compute the Shapley value attributions for each word in the prompt as discussed in section \ref{sec:theory}. 
On GPT-4o we get the following scores.

\begin{figure}[H]
\centering
\begin{tikzpicture}
\begin{axis}[
    width=15cm, height=4cm,
    ybar,
    symbolic x coords={which, Stock, do, you, invest, in, if, stock, B, makes, a, profit, 70\%, of, the, time},
    xtick=data,
    x tick label style={rotate=45, anchor=east},  
    ymin=-0.05, ymax=0.4,
    ytick={-0.05, 0, 0.1, 0.2, 0.3, 0.4},  
    scaled y ticks=false,  
    yticklabel style={/pgf/number format/fixed},  
    xlabel={Words},
    ylabel={Influence scores},
    bar width=15pt
]
\addplot coordinates {
    (which, -0.00568105)
    (Stock, 0.00519236)
    (do, 0.0150998)
    (you, 0.01750905)
    (invest, 0.09858772)
    (in, 0.02046272)
    (if, 0.06535568)
    (stock, -0.0226652)
    (B, 0.36325925)
    (makes, 0.06793707)
    (a, 0.0524025)
    (profit, 0.09271258)
    (70\%, 0.13876917)
    (of, 0.0145994)
    (the, 0.01747991)
    (time, 0.03700515)
};
\end{axis}
\end{tikzpicture}
\caption{Shapley score attribution for positively framed prompt}
\end{figure}
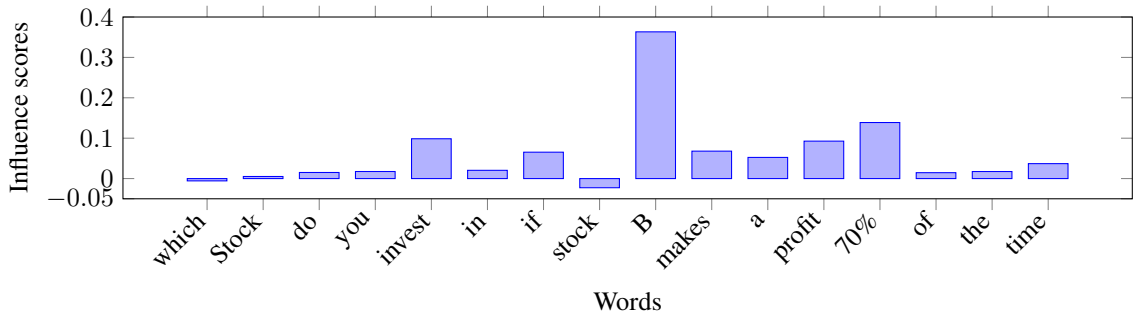

As can be noted, maximum contribution to preferring stock B can be attributed to the words "B" and "70\%", while the word "profit" has a moderate contribution to improving the prediction in favor of stock B. 
Most of the other words have little to no contribution in improving the overall score.
Hence, it can be reasoned, the response is quite positively induced due to a high success rate of 70\%.

Contemplating on this graph, it's hard to escape the similarity it presents with the attention map when considered for a single token. In a sense, Shapley score depiction can be seen as an attention map to the underlying cognition process that leads to stock B being the favoured choice, with more significant words receiving a higher influence score in the model decision making process.

\begin{figure}[H]
\centering
\begin{tikzpicture}
\begin{axis}[
        width=15cm, height=4cm,
    ybar,
    symbolic x coords={which, Stock, do, you, invest, in, if, stock, B, makes, a, loss, 30\%, of, the, time},
    xtick=data,
    x tick label style={rotate=45, anchor=east},  
    ymin=-0.03, ymax=0.03,  
    ytick={-0.03, -0.02, -0.01, 0, 0.01, 0.02, 0.03},  
    xlabel={Words},
    ylabel={Influence scores},
    bar width=15pt
]
\addplot coordinates {
    (which, -0.01527253)
    (Stock, -0.01427887)
    (do, 0.01394382)
    (you, 0.01706567)
    (invest, 0.02695871)
    (in, 0.00579732)
    (if, 0.00228201)
    (stock, -0.00110056)
    (B, -0.00811044)
    (makes, -0.00291399)
    (a, 0.01540873)
    (loss, -0.02658258)
    (30\%, 0.02137472)
    (of, -0.00300819)
    (the, 0.00715951)
    (time, -0.02039802)
};
\end{axis}
\end{tikzpicture}
\caption{Shapley score attribution for negatively framed prompt}
\end{figure}
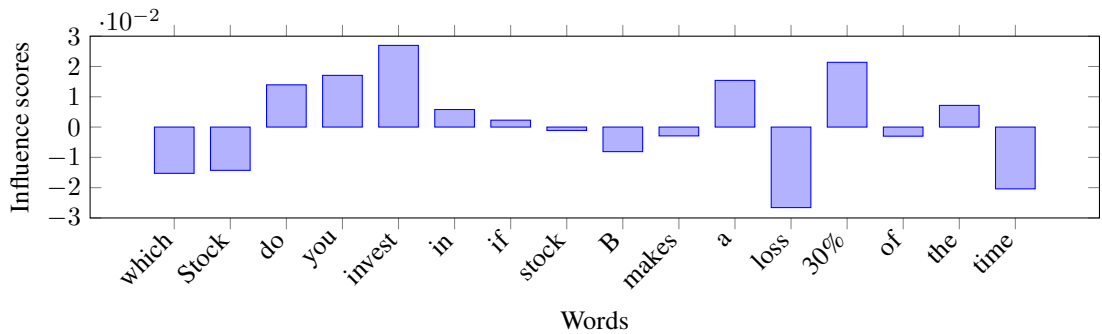

Revisiting the graph \ref{fig:framing_effect_prob}, the model's prediction shifts against B as preferred stock when framed in a negative manner, encouraging us to compare the differences in sentences that lead to a behaviour change.
Plotting the Shapley value graph for modified prompt with "B" as observed token, the word "loss" has the highest absolute contribution to the prediction. With a highly negative score it biases the model severely against stock B as the preferred choice, shaping it's decision process in a stark contrast to the positively framed example.
This change in it's reasoning is a classical example of framing effect where similar ideas presented in 2 different manners, elucidate two different outputs from a respondent.

Building further, we start looking for a point where the model finally decodes a low loss percentage as a positive signal. We label this as  model's cognitive bias barrier, namely, the point at which an LLM is able to understand a lower loss percentage corresponds to a higher profit percentage and associates it as a positive factor.

\begin{tcolorbox}[
    colback=white,
    colframe=black,
    boxrule=1pt,
    arc=5mm, 
    left=10pt,right=10pt,top=10pt,bottom=10pt,
    boxsep=0pt,
    width=\textwidth
]

\textbf{\textit{Prompt A}}: Given two stocks A and B, which stock do you invest in if stock A makes a loss i\% of the time?\\
\textbf{\textit{Prompt B}}: Given two stocks A and B, which stock do you invest in if stock B makes a loss i\% of the time?
\end{tcolorbox}

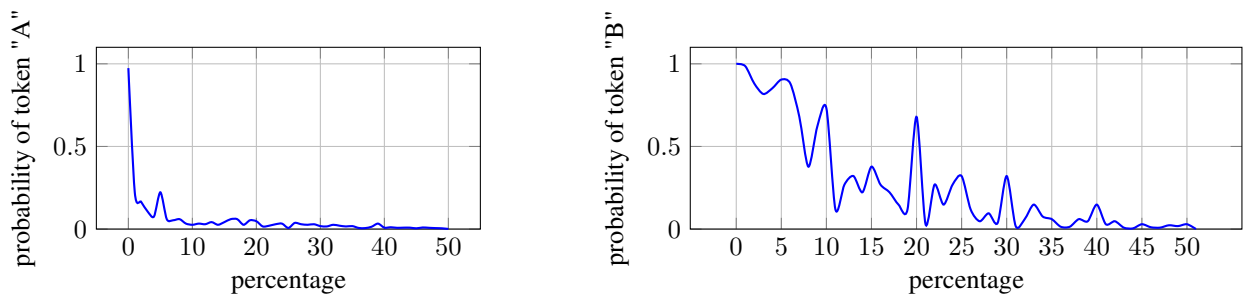
\begin{figure}[h!]
    \centering
    \begin{subfigure}{0.45\textwidth}
        \centering
        \begin{tikzpicture}
        \begin{axis}[
            xlabel={percentage},
            ylabel={probability of token "A"},
            grid=major,
            width=0.9\textwidth,
            height=4cm,
            ymin=0, ymax=1.1, 
            xtick={0,10,...,50}, 
            smooth,
        ]
        \addplot[blue, thick] coordinates {
            (0, 0.9740426358351159)
            (1, 0.2227001308356713)
            (2, 0.16451646876541565)
            (3, 0.10669059737686205)
            (4, 0.07585817915653906)
            (5, 0.2227001423705874)
            (6, 0.06008666508875193)
            (7, 0.05340331133658865)
            (8, 0.06008665659825566)
            (9, 0.03308597102312977)
            (10, 0.02595735053544098)
            (11, 0.03308598309142318)
            (12, 0.029312232847598187)
            (13, 0.04208769898480224)
            (14, 0.025957360202243877)
            (15, 0.04208772155621078)
            (16, 0.06008662530483549)
            (17, 0.060086655426321384)
            (18, 0.025957350001260895)
            (19, 0.053403313257217366)
            (20, 0.0474258780171163)
            (21, 0.015906389615792976)
            (22, 0.020332346674071015)
            (23, 0.029312207424965174)
            (24, 0.03308598488801921)
            (25, 0.006692846589437779)
            (26, 0.03732690175997017)
            (27, 0.0293122352864204)
            (28, 0.025957351389207895)
            (29, 0.02931222918658003)
            (30, 0.017986203610006343)
            (31, 0.015906389166090336)
            (32, 0.025957344680271017)
            (33, 0.020332358234719603)
            (34, 0.015906395147835924)
            (35, 0.017986195429461078)
            (36, 0.006692848996426002)
            (37, 0.005911067290038843)
            (38, 0.014063622052127592)
            (39, 0.03308597763539559)
            (40, 0.008577482409297656)
            (41, 0.01098694545119765)
            (42, 0.007577241593614325)
            (43, 0.008577486133034884)
            (44, 0.008577489396493727)
            (45, 0.00460957153795823)
            (46, 0.00970847389967986)
            (47, 0.007577240345066756)  
            (48, 0.005911070991523906)
            (49, 0.004609570878593534)
            (50, 0.0008040860707718326)
        };
        \end{axis}
        \end{tikzpicture}
        \caption{Probability scores for varying loss percentages - GPT-4o}
        \label{fig:lossAgpt4o}
    \end{subfigure}
    \hspace{0.05\textwidth} 
    \begin{subfigure}{0.45\textwidth}
        \centering
        \begin{tikzpicture}
        \begin{axis}[
            xlabel={percentage},
            ylabel={probability of token "B"},
            grid=major,
            width=1.2\textwidth,
            height=4cm,
            ymin=0, ymax=1.1, 
            xtick={0,5,...,50}, 
            smooth,
        ]
        \addplot[blue, thick] coordinates {
            (0, 9.99796573e-01) (1, 9.85936371e-01) (2, 8.80797080e-01) (3, 8.17574479e-01)
            (4, 8.51952797e-01) (5, 9.04650545e-01) (6, 8.80797085e-01) (7, 6.79178714e-01)
            (8, 3.77540678e-01) (9, 6.22459331e-01) (10, 7.31058578e-01) (11, 1.19202911e-01)
            (12, 2.68941430e-01) (13, 3.20821301e-01) (14, 2.22700147e-01) (15, 3.77540674e-01)
            (16, 2.68941410e-01) (17, 2.22700152e-01) (18, 1.48047195e-01) (19, 1.19202916e-01)
            (20, 6.79178690e-01) (21, 2.93122321e-02) (22, 2.68941422e-01) (23, 1.48047200e-01)
            (24, 2.68941433e-01) (25, 3.20821290e-01) (26, 1.19202922e-01) (27, 4.74258760e-02)
            (28, 9.53494567e-02) (29, 3.73268891e-02) (30, 3.20821305e-01) (31, 1.79862124e-02)
            (32, 6.00866419e-02) (33, 1.48047194e-01) (34, 7.58581727e-02) (35, 6.00866480e-02)
            (36, 1.40636225e-02) (37, 1.40636258e-02) (38, 6.00866612e-02) (39, 4.74258777e-02)
            (40, 1.48047197e-01) (41, 2.93122309e-02) (42, 4.74258739e-02) (43, 8.57748650e-03)
            (44, 4.07013791e-03) (45, 2.93122338e-02) (46, 1.09869405e-02) (47, 8.57748683e-03)
            (48, 2.29773689e-02) (49, 1.79862079e-02) (50, 2.93122317e-02) (51, 1.58436296e-04)
        };

        \end{axis}
        \end{tikzpicture}
        \caption{Probability scores for varying loss percentages - GPT-4o-mini}
        \label{fig:lossAgpt4omini}
    \end{subfigure}
    \caption{Comparison of probability scores for tokens "A" and "B" at varying loss percentages.}
\end{figure}

We evaluate prompt A on GPT-4o-mini and prompt B on GPT-4o to compare the differences in cognitive bias barrier in both models. 
While the graph is straightforward for GPT-4o, where for each value greater than 0\% the model consistently displays "A" as preferred output, the graph is a bit more jagged when tested on GPT-4o-mini.\\
Surprisingly in GPT-4o-mini, the model upholds it's barrier for higher values of failure rates as well, maintaining it up to loss percentages of 20\%, indicative by a greater preference for stock B over A in it's graph. 
A higher cognitive bias barrier reflects the model's comprehension of lower loss rates' desirability and indicates enhanced robustness in its decision-making process.

\subsection{Number bias}
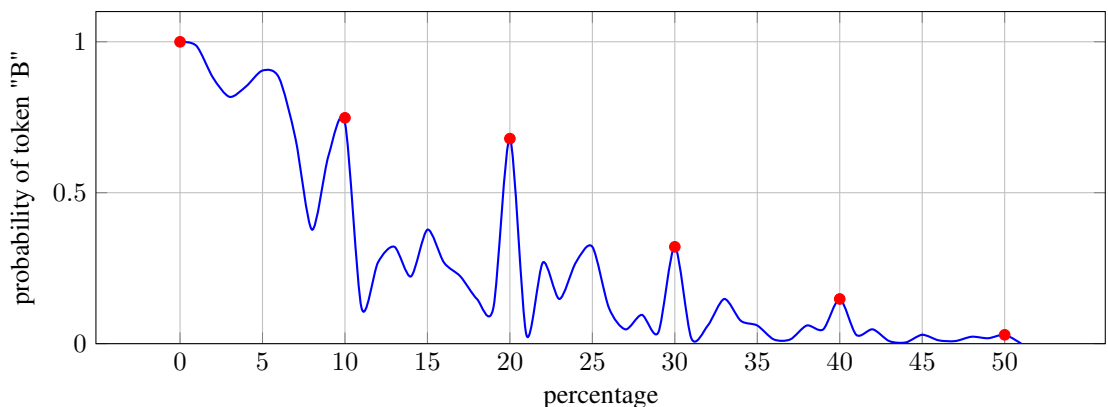
\begin{figure}[h]
    \centering
    \begin{tikzpicture}
    \begin{axis}[
        xlabel={percentage},
        ylabel={probability of token "B"},
        grid=major,
        width=15cm, 
        height=6cm, 
        ymin=0, ymax=1.1, 
        xtick={0,5,...,50}, 
        smooth, 
        samples=200, 
    ]

    \addplot[blue, thick] coordinates {
        (0, 9.99796573e-01) (1, 9.85936371e-01) (2, 8.80797080e-01) (3, 8.17574479e-01)
        (4, 8.51952797e-01) (5, 9.04650545e-01) (6, 8.80797085e-01) (7, 6.79178714e-01)
        (8, 3.77540678e-01) (9, 6.22459331e-01) (10, 7.31058578e-01) (11, 1.19202911e-01)
        (12, 2.68941430e-01) (13, 3.20821301e-01) (14, 2.22700147e-01) (15, 3.77540674e-01)
        (16, 2.68941410e-01) (17, 2.22700152e-01) (18, 1.48047195e-01) (19, 1.19202916e-01)
        (20, 6.79178690e-01) (21, 2.93122321e-02) (22, 2.68941422e-01) (23, 1.48047200e-01)
        (24, 2.68941433e-01) (25, 3.20821290e-01) (26, 1.19202922e-01) (27, 4.74258760e-02)
        (28, 9.53494567e-02) (29, 3.73268891e-02) (30, 3.20821305e-01) (31, 1.79862124e-02)
        (32, 6.00866419e-02) (33, 1.48047194e-01) (34, 7.58581727e-02) (35, 6.00866480e-02)
        (36, 1.40636225e-02) (37, 1.40636258e-02) (38, 6.00866612e-02) (39, 4.74258777e-02)
        (40, 1.48047197e-01) (41, 2.93122309e-02) (42, 4.74258739e-02) (43, 8.57748650e-03)
        (44, 4.07013791e-03) (45, 2.93122338e-02) (46, 1.09869405e-02) (47, 8.57748683e-03)
        (48, 2.29773689e-02) (49, 1.79862079e-02) (50, 2.93122317e-02) (51, 1.58436296e-04)
    };

    \addplot[
        only marks,
        mark=*,
        mark size=2pt,
        mark options={red, fill=red}
    ] coordinates {
        (0, 9.99796573e-01) (10, 7.48058578e-01) (20, 6.79178690e-01) 
        (30, 3.20821305e-01) (40, 1.48047197e-01) (50, 2.93122317e-02)
    };
    \end{axis}
    \end{tikzpicture}
    \caption{Probability scores for prompt B on GPT-4o-mini.}
    \label{fig:token-probability}
\end{figure}

Examining the prompt used in section~\ref{sec:frame}, the jaggedness in graph for varying values of loss percentages compelled us to question whether any patterns emerge within the irregularities. We consider one of the  prompts that exhibits a pattern even in the unevenness observed in the figure - \ref{fig:lossAgpt4omini} 
The given graph plots the changes in LLM output probability scores for stock A as preferred option when considered for all percentages in range [0,100].
For varying values of percentages, percentage values which are multiple of 10 seem to be the local maxima in a range of (+10,-10), where the graph exhibits very prominent peaks. These spikes in probability scores indicate a particular bias in the graph going against a consistent expected decrease that should be demonstrated for increasing values of loss percentages.
Such strong preference for values in multiples of 10 is evident of a round number bias in model behaviour. 

We further study round number bias in LLMs by analyzing their performance on a dataset of  student essays. For each essay, the model is prompted to assign a grade between 1 to 100. The below graph shows the output score distribution for 1000 essays.

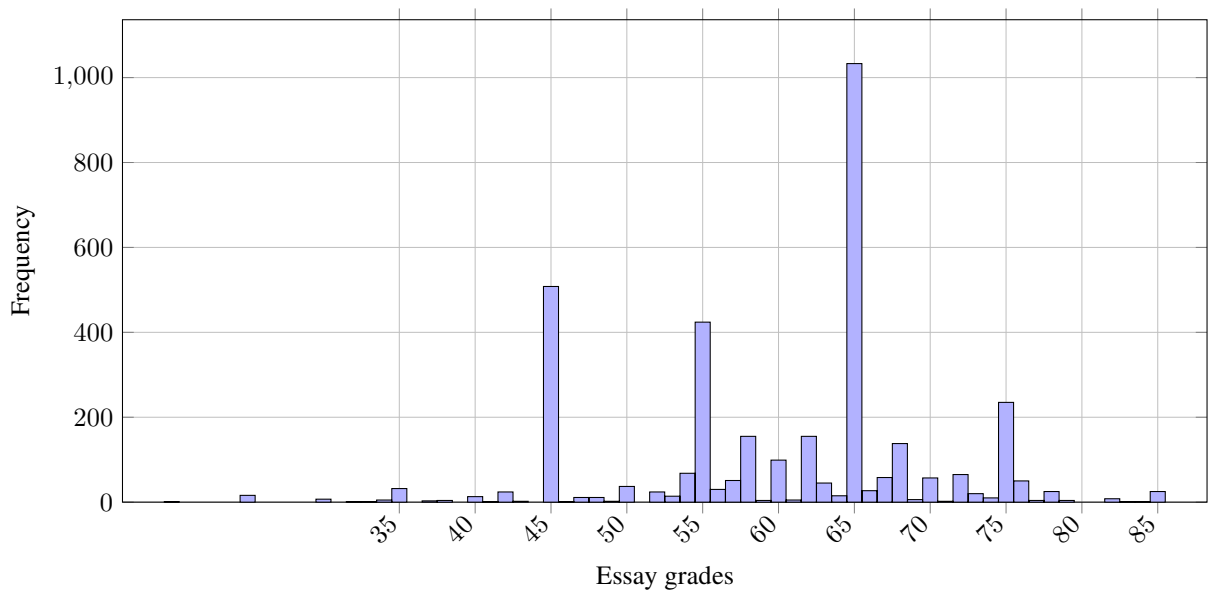
\begin{tikzpicture}
    \begin{axis}[
        ybar,
        bar width=0.2cm,
        width=16cm,
        height=8cm,
        xlabel={Essay grades},
        ylabel={Frequency},
        xtick={35,40,45,50,55,60,65,70,75,80,85},
        xticklabel style={rotate=45, anchor=east},
        ymin=0,
        grid=both,
        major grid style={line width=0.2pt,draw=gray!50},
        minor grid style={line width=0.1pt,draw=gray!20},
        nodes near coords align={vertical},
        enlarge x limits=0.05
    ]
        \addplot[fill=blue!30,draw=black] coordinates {
            (20,1) (25,16) (30,7) (32,1) (33,1) (34,5) (35,32) (37,3) (38,4) 
        (40,13) (41,1) (42,24) (43,2) (45,508) (46,1) (47,11) (48,11) 
        (49,2) (50,37) (52,24) (53,14) (54,68) (55,424) (56,30) (57,51) 
        (58,155) (59,4) (60,99) (61,5) (62,155) (63,45) (64,15) (65,1033) 
        (66,27) (67,58) (68,138) (69,6) (70,57) (71,2) (72,65) (73,20) 
        (74,10) (75,235) (76,50) (77,4) (78,25) (79,4) (82,8) (83,1) 
        (84,1) (85,25)
        };
    \end{axis}
\end{tikzpicture}

As illustrated in the graph, the model exhibits notable preference in assigning grades that are in multiples of 5, displaying a significant round number bias.
Interestingly, in all of our experiments across varying datasets and metrics, models consistently showed the strongest preference for assigning a score of 65 out of 100.

\subsection{Anchoring Effect} \label{sec:anchoring_effect}

Defined as a cognitive bias that describes how people tend to rely too heavily on the first piece of information they hear when making decisions, \textit{anchoring effect} was first introduced by Kahnemann and Tversky in their paper \cite{tversky1974judgment}.  During decision making, anchoring occurs when individuals use an initial piece of information to make subsequent judgments.
This initial piece of information known as the "anchor", is responsible for instilling a bias towards the respondent's answer.\\
To understand this presence in LLMs we analysed frontier LLM responses on two  different prompts, providing a varied perspective on the same underlying concept of uncovering an anchoring bias.
\begin{tcolorbox}[
    colback=white,
    colframe=black,
    boxrule=1pt,
    arc=5mm, 
    left=10pt,right=10pt,top=10pt,bottom=10pt,
    boxsep=0pt,
    width=\textwidth
]
\textit{Anchoring Prompt:}\\
Estimate the number of jellybeans in a hidden jar. Your friend guesses 750. What's your estimate? \\
\hspace{1em}A) 50 
\hspace{1em}B) 200 
\hspace{1em}C) 800 
\hspace{1em}D) 1200

\end{tcolorbox}

Despite prompting the model of the fact that the friend "guesses" their answer, the model seems to be quite biased in picking option C - 800, the closest value to the 750. 
This presence of consistent bias prompts us to gauge the Shapley score to determine whether the bias can be attributed to the presence of an "anchor token".\\

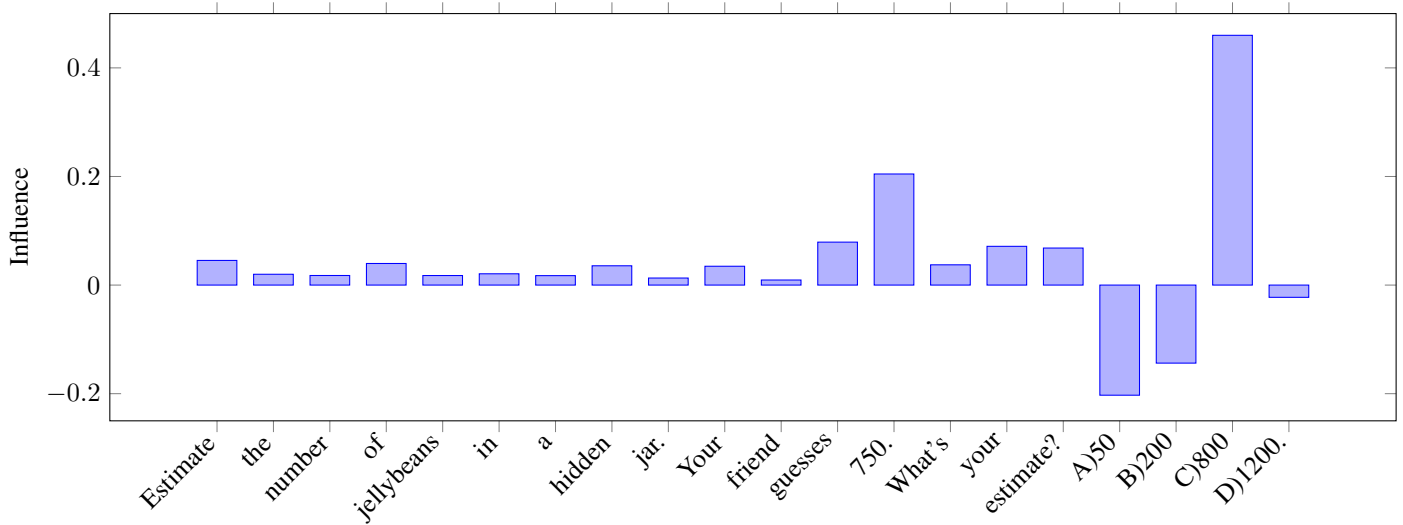
\begin{figure}[!h]
    \centering
    \begin{tikzpicture}
    \begin{axis}[
        ybar,
        bar width=15pt,  
        ylabel={Influence},  
        symbolic x coords={Estimate, the, number, of, jellybeans, in, a, hidden, jar., Your, friend, guesses, 750., What's, your, estimate?, A)50, B)200, C)800, D)1200.},  
        xtick=data,
        x tick label style={rotate=45, anchor=east},  
        ymin=-0.25,  
        ymax=0.5,  
        width=15cm,  
        height=7cm,  
        x=0.75cm,  
    ]
    \addplot coordinates {
        (Estimate, 0.04539333377940747) 
        (the, 0.019921455674183276) 
        (number, 0.01764864903151611) 
        (of, 0.039769246021426785) 
        (jellybeans, 0.017582229880891368) 
        (in, 0.020793353710298668) 
        (a, 0.017395586378404722) 
        (hidden, 0.03554080541029437) 
        (jar., 0.01297868398202977) 
        (Your, 0.03468530571089872) 
        (friend, 0.009336178751951588) 
        (guesses, 0.07909492411964347) 
        (750., 0.20458324934813463) 
        (What's, 0.0373110013354196) 
        (your, 0.07136958624555267) 
        (estimate?, 0.06820421651701587) 
        (A)50, -0.2028079798885301) 
        (B)200, -0.14371885420387992) 
        (C)800, 0.45998779341448437) 
        (D)1200., -0.022555052931253216)
    };
    \end{axis}
    \end{tikzpicture}
    \caption{Shapley score analysis on anchoring prompt}
\end{figure}
Noting the contribution of each token in output probability, the highest contributions for prediction lie in C)800, scoring quite high understandably due to being the preferred choice by LLM.  
Interestingly the token 750 possesses the second highest Shapley attribution, influencing the output probability in a significant manner, thus displaying an evident anchoring effect in the model behaviour.
The similarity of this shapley score attribution to a transformer model's attention map is hard to miss. We hypothesise, in a manner, shapley score attribution to tokens is like the attention map to a model's cognition process, providing us with a rough schema of  most important tokens/contributors in it's underlying decision making process. \\
Building upon an established presence of anchoring effect in the LLM, we test our model on the infamous letter counting problems that LLMs are so unreliable on.
We contrast the output of 2 different prompts, that together provide strong evidence of an LLMs anchoring bias.

\begin{tcolorbox}[
    colback=white,
    colframe=black,
    boxrule=1pt,
    arc=5mm, 
    left=10pt,right=10pt,top=10pt,bottom=10pt,
    boxsep=0pt,
    width=\textwidth
]
\textit{Original prompt:}
How many letters "r" are there in the word barrier?\\
\textit{Persuasion prompt:} How many letters "r" are there in the word barrier? Your friend insists it's 4.
\end{tcolorbox}

Every frontier LLM fails on the original prompt and provides an incorrect count of 2.
The correct count being 3 of course.
We perform Shapley value analysis on both prompts with 3 as the prediction token this time.

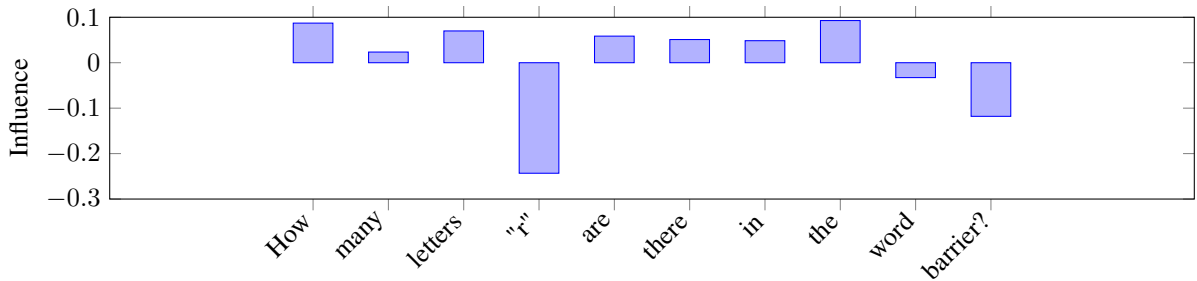
\begin{figure}[H]
    \centering
    \begin{tikzpicture}
    \begin{axis}[
        ybar,
        bar width=15pt,  
        ylabel={Influence},  
        symbolic x coords={How, many, letters, "r", are, there, in, the, word, barrier?},  
        xtick=data,
        x tick label style={rotate=45, anchor=east},  
        ymin=-0.3,  
        ymax=0.1,  
        width=16cm,  
        height=4cm,  
        enlarge x limits=0.3,  
        trim axis left,
        trim axis right
    ]
    \addplot coordinates {
        (How, 0.08714158317764613) 
        (many, 0.023463473812845647) 
        (letters, 0.06991388245404374) 
        ("r", -0.24321136450571407) 
        (are, 0.058502029043638056) 
        (there, 0.05101058235840297) 
        (in, 0.04846258271549338) 
        (the, 0.0927852556471787) 
        (word, -0.0327482403812571) 
        (barrier?, -0.11799289615167535)
    };
    \end{axis}
    \end{tikzpicture}
    \caption{Bar graph of influence values for original prompt.}
\end{figure}
As can be noted from graph, the word barrier has a significant negative contribution to the expected response of 3, corroborating the model's incorrect answer. 
\begin{figure}[H]
    \centering
    \begin{tikzpicture}
    \begin{axis}[
        ybar,
        bar width=15pt,  
        ylabel={Influence},  
        symbolic x coords={How, many, letters, "r", are, there, in, the, word, barrier, Your, friend, insists, it's, 4},  
        xtick=data,
        x tick label style={rotate=45, anchor=east},  
        ymin=-0.1,  
        ymax=0.25,  
        width=16cm,  
        height=4.4cm,  
        enlarge x limits=0.2,  
    ]
    \addplot coordinates {
        (How, 0.0680909) 
        (many, 0.07852322) 
        (letters, 0.02437296) 
        ("r", 0.22323869) 
        (are, -0.00798833) 
        (there, 0.00994738) 
        (in, 0.0038298) 
        (the, 0.03992313) 
        (word, -0.05308434) 
        (barrier, 0.05592742) 
        (Your, 0.061466) 
        (friend, 0.03166812) 
        (insists, 0.05633916) 
        (it's, 0.04901747) 
        (4, 0.10678091)
    };
    \end{axis}
    \end{tikzpicture}
    \caption{Bar graph of influence values for persuasion prompt}
\end{figure}
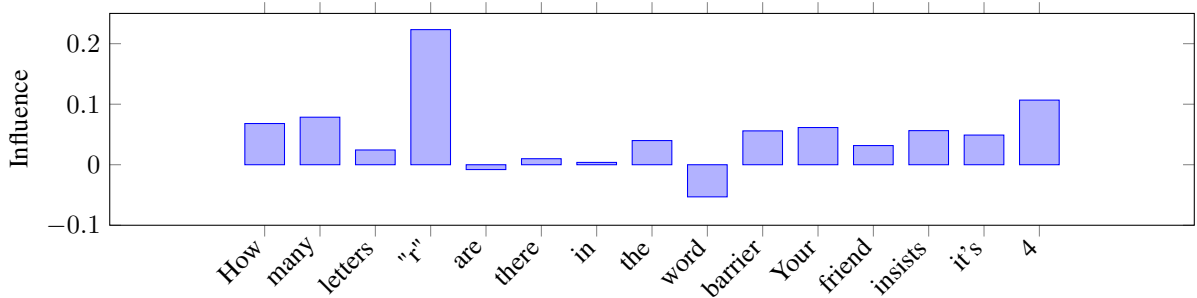

Interestingly, both Llama 3.1 and GPT-4o provide the correct response of 3 on the persuasion prompt , despite being provided an incorrect count of 4 as input, being suggestive of an underlying anchoring effect at play in prodding the model output to the correct response.
Noting from the graph, the token 4 has the second highest shapley value contribution after "r", establishing it's significance as the anchor token, and confirming it's anchoring effect in influencing the response.
We hypothesize that being provided with an initial incorrect count of the letter "r" prompts the model to reflect on the count and come up with the correct answer. 

Anchoring effect is particularly noticeable in LLMs due to their significant in-context learning abilities\cite{olsson2022context}, the ability to pick up patterns in the input prompts, which eventually shows up as "anchor" on these tokens, in the model's reasoning process.

If a prompt contains several high numerical values, the model might continue suggesting similar high values in subsequent outputs, reflecting an anchoring bias. Conversely, if the initial numbers are low, the responses may lean towards lower figures.

\subsection{Representativeness Heuristic}
\label{sec:repr_heuristic}
Representativeness heuristic can manifest itself due to knowledge gaps in people's thinking where they don't consider choices and alternatives that aren't as ubiquitous.
In LLMs this knowledge gap can exist due to an incompleteness or inconsistency in internal world model acquired through various pre-training and post-training processes \cite{ouyang2022training}.

Representativeness heuristic in LLMs can show up due to numerous reasons, such as a skewness in the data distribution the model is trained on, causing it to output in-distribution data more than logically and critically accurate responses. To test for this bias we use 2 different prompts, one being derivative of a popular example used to elucidate heuristic biases in humans. 
\begin{tcolorbox}[
    colback=white,
    colframe=black,
    boxrule=1pt,
    arc=5mm, 
    left=10pt,right=10pt,top=10pt,bottom=10pt,
    boxsep=0pt,
    width=\textwidth
]
\textit{Representativeness Heuristic Prompt:}\\
Mahesh is an extremely smart guy who's great at mental arithmetic. Is he more likely to be a cop or a Mathematics field medalist?

\end{tcolorbox}

Despite Mahesh having strong arithmetic skills which is more likely to be a trait in a fields medallist than a cop, the incomparably high ratio of cops compared to field's medallists sufficiently favors Mahesh to being a cop over being a field's medallist.\\
Most frontier LLMs provide an incorrect response, being biased by a strong arithmetic acumen in Mahesh and not considering the prior instilled due to a much higher number of cops than fields medallists.\\
Most frontier LLMs display a strong proclivity to falling prey to representativeness heuristic bias,  with GPT-4o being one of the models successfully able to understand this as probing for base rate fallacy bias and gives the correct response. However, it's smaller version GPT-4o-mini still provides an unsuccessful response marking the differences in understanding and reasoning capabilities between these models.

We use 2 additional prompts to provide a different perspective into how representativeness heuristic may show up in a model's logical reasoning process.

\begin{tcolorbox}[
    colback=white,
    colframe=black,
    boxrule=1pt,
    arc=5mm, 
    left=10pt,right=10pt,top=10pt,bottom=10pt,
    boxsep=0pt,
    width=\textwidth
]
\textit{Monty Hall problem:}\\
\textbf{Prompt A}: 
You are a part of a game show, where the host Monty shows you three doors, behind two is a car, and behind one nothing. You have to pick a door as your preference in order to win the presumed car behind it. Initially you pick door number 1. The host opens door 2 and shows it's empty. Now he offers you the choice to switch to door 3. Do you take that choice or stick with your current option?\\
\textbf{Prompt B}: 
The famous gameshow host Monty Hall has decided to play a game with me. The game consists of 3 doors, behind two of those doors is a goat and behind one is an extremely poisonous snake that can kill humans. He tells me to pick a door, I pick door 1. Now Monty opens door number 3 and reveals a goat behind it. He offers me the choice to stick with my current option or switch to door number 2, what should I do?

\end{tcolorbox}

With the obvious expectation on frontier models having the famous Monty hall problem in it's training data in one form or the other, we evaluate the model responses on slight variations of the problem. 
On both prompts, GPT-4o provides a clearly incorrect response and reasoning to the decision making involved in the game. Despite there not being any similarity in the logical reasoning involved in prompt A and the Monty hall problem, due to an existing mental heuristic to think along the lines of the Monty hall problem, the model presents an incorrect reasoning process. 
Similarly on prompt B, the most simple thinking of avoiding the door with snake behind it evades the model due to it's inherent priors to reason along the lines of the Monty Hall problem. 

These 2 coupled with other prompts structured in a manner similar as the Monty Hall problem provide a strong evidence for a mental heuristic in frontier LLMs (such as GPT-4o) to be biased to mimic data most frequented in it's training process, despite not being supportive of a sound reasoning process. \\
Deciphering the presence of representativeness heuristic bias in LLMs questions the very notion of  "reasoning" in language models and whether they are capable of actual cognition or are simply mimicking frequented patterns on a superficial level.

\subsection{Priming effect}
\label{sec:priming}
The priming effect refers to a psychological phenomenon where exposure to one stimulus influences the response to a subsequent stimulus, often without the subject's awareness of this influence. \\
This effect occurs when prior exposure to certain information, or stimuli unconsciously affects how the subject perceives or responds to later information. Essentially, the initial stimulus "primes" or prepares the mind to perceive related information in a specific manner.

To detect this effect we instill a bias in our prompt where an initial conditioning on a particular colour (prior stimuli) influences posterior decision making in LLM.

\begin{tcolorbox}[
    colback=white,
    colframe=black,
    boxrule=1pt,
    arc=5mm, 
    left=10pt,right=10pt,top=2pt,bottom=2pt,
    boxsep=0pt,
    width=\textwidth
]
\textit{Priming effect prompt:}\\
My friend arrived to the fruit market wearing his favourite bright red coloured shirt. Which of these fruits is he most likely to buy?
\hspace{1em}A) Banana 
\hspace{1em}B) Apple 
\hspace{1em}C) Pear 
\hspace{1em}D) Grapes 
\end{tcolorbox}
Plotting the preference graph for each of the options it's clear to see the colour exposure biasing the model significantly to pick Apple as it's preferred response, a promising indication of potential priming effect in play. 

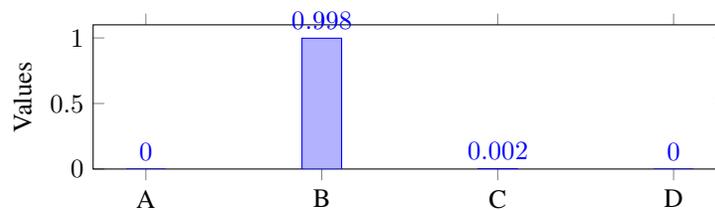
\begin{figure}[!h]
    \centering
    \begin{tikzpicture}
        \begin{axis}[
            ybar,
            symbolic x coords={A, B, C, D},
            xtick=data,
            ymin=0, ymax=1.1,
            ylabel={Values},
            nodes near coords,
            every node near coord/.append style={/pgf/number format/fixed, /pgf/number format/precision=4},
            bar width=15pt,
            width=10cm,
            height=3.5cm
        ]
        \addplot coordinates {(A, 0.0) (B, 0.998) (C, 0.002) (D, 0.0)};
        \end{axis}
    \end{tikzpicture}
    \caption{Preference scores for each option, normalized to total score of 1.}
\end{figure}

We further explore if this bias is a consequence of the colour stimulus by plotting the Shapley value graph for our prompt in a manner similar to section~\ref{sec:anchoring_effect}.
\noindent
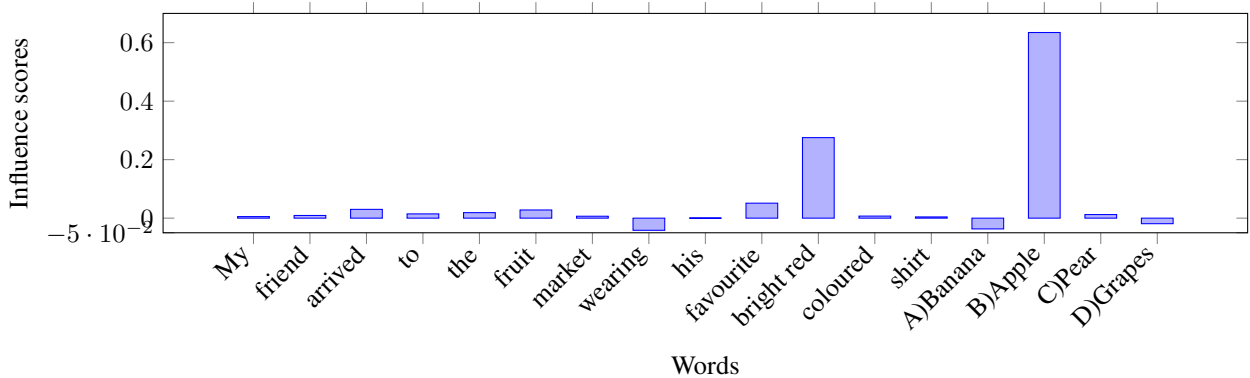
\begin{figure}[H]
\centering
\begin{tikzpicture}
\begin{axis}[
    width=16cm, height=4.5cm,  
    ybar,
    symbolic x coords={My, friend, arrived, to, the, fruit, market, wearing, his, favourite, bright red, coloured, shirt, A)Banana, B)Apple, C)Pear, D)Grapes},
    xtick=data,
    x tick label style={rotate=45, anchor=east},  
    ymin=-0.05, ymax=0.7,  
    ytick={-0.05, 0, 0.2, 0.4, 0.6},  
    xlabel={Words},
    ylabel={Influence scores},
    bar width=12pt
]
\addplot coordinates {
    (My, 0.005306822676365995)
    (friend, 0.009154325918908253)
    (arrived, 0.03005195754692578)
    (to, 0.014561499150224803)
    (the, 0.018830342059046688)
    (fruit, 0.027939648242359446)
    (market, 0.006569712799756442)
    (wearing, -0.04155613595563828)
    (his, 0.00016092224788036194)
    (favourite, 0.051186217618485186)
    (bright red, 0.27533397755099465)
    (coloured, 0.007032322962659264)
    (shirt, 0.004196766759901629)
    (A)Banana, -0.03681348535806377)
    (B)Apple, 0.6347359042451243)
    (C)Pear, 0.012271542085091784)
    (D)Grapes, -0.018962359844051164)
};
\end{axis}
\end{tikzpicture}
\caption{Shapley value attribution graph}
\end{figure}
As can be noted from the influence scores, the model output is significantly favoured by the initial trigger (bright red) in determining the preferred response as apples. This is a substantial indicator of priming effect where subsequent decisions are strongly affected by a prior stimuli.

\section{Discussion \& Conclusion}
\label{sec:discuss}

\begin{table}[htbp]
\centering
\caption{\textit{Cognitive Bias Evaluation}}
\label{tab:llm-evaluations}
\resizebox{\textwidth}{!}{%
\begin{tabular}{|p{0.30\textwidth}|p{0.10\textwidth}|p{0.10\textwidth}|p{0.10\textwidth}|p{0.10\textwidth}|p{0.10\textwidth}|p{0.10\textwidth}|}
\hline
\textbf{Cognitive Bias} & \textbf{GPT-4o} & \textbf{GPT-4o-mini} & \textbf{Claude 3.5 Sonnet} & \textbf{Llama 3 405B} & \textbf{Google Gemini} & \textbf{Mistral Large} \\
\hline
Framing effect & \ding{51} & \ding{51} & \textbf{-} & \ding{51} & \textbf{-} & \textbf{-} \\
\hline
Anchoring effect (Jellybeans prompt) & \ding{51} & \ding{51} & \ding{51} & \ding{51} & \ding{51} & \ding{51} \\
\hline
Representativeness heuristic (Monty Hall prompt) & \ding{51} & \ding{51} & \ding{55} & \ding{51} & \ding{51} & \ding{51} \\
\hline
Priming effect & \ding{51} & \ding{51} & \ding{51} & \ding{51} & \ding{51} & \ding{51} \\
\hline
\end{tabular}%
}
{\small \textit{Note:} A \ding{51}  indicates presence of bias, a \ding{55} indicates its absence, and \textbf{-} indicates refusal to respond.}
\end{table}

Our experiments have helped us interpret cognitive shortcomings instilled in LLMs, which arise from training heuristics such as tokenization \ref{sec:anchoring_effect} and model characteristic such as in-context learning \ref{sec:priming}, while also reflecting biases prevalent in LLMs due to the training data it reflects \ref{sec:frame}. Evaluations on representativeness heuristic \ref{sec:repr_heuristic} also reveal that slightest deviations from a standard archetype, not in the model's data, can cause the model to fall prey to the same cognitive bias traps as humans. 
Furthermore, we present evidence of round number bias in Language models such as GPT-4o-mini,

Despite the clear similarities in our experimentation involving anchoring \ref{sec:anchoring_effect} and priming effect \ref{sec:priming}  as both involve prior information influencing subsequent judgments, there are subtle differences to the manner in which this influence is incorporated. While anchoring relies on a strong dependence on an initial estimate or value to make downstream decisions, priming effect rely on associations developed in a subject's (system) world model and internal representations of concepts.\\
Although, both are incorporated due to the attention mechanism,a key component responsible for trading information between tokens \cite{vaswani2017attention}, in a GPT-based model priming effect relies more on the associated mappings developed in the model, a characteristic particular to the model's MLP layers  \cite{geva2020transformer}, while the anchoring mechanism relies on the exact contents of the anchoring token, and hence is a much more direct influence.

\section{Limitations}
Much of our work relies of the output probability distribution produced by an LLM. While significant prior work demonstrates model responses to change due to token noise rather than actual reasoning \cite{mohammadi2024wait}, most of our prompts have extremely strong preferences for one choice over the other, as can be viewed in figure \ref{fig:stock-comparison}, so variances in sentences or switching the order of options doesn't affect the probability outputs by much. \\
In evaluations such as round number bias where we explicitly plot the probability distribution over varying values of percentages, while changing the prompt did alter the graph, the points of interest (i.e, mutiples of 5s) displayed a consistent pattern of being local maxima and were a consistent theme of interest for us to inspect. \\
In order to get the output in correct format we depend on system prompts to guide the model to the expected output format. Different system prompts produced slightly different probability distributions but much of the output score for token of interest was maintained across various system prompts we tried.\\
LLMs such as Mistral Large, Claude and google Gemini don't provide APIs that let us access the probability score for tokens. For such models we could assess only for the presence of tested biases through subjective evaluation. However, for GPT-4o, GPT-4o-mini, and Llama-3 we were able to confirm our outputs and the underlying reasoning behind various cognitive biases through Shapley value computation graphs.
Most of our experimentation was conducted prior to release of GPT-o1 model and we haven't included it in any of our experiments or analysis.
\bibliographystyle{unsrt}  
\bibliography{references}  

\end{document}